\pdfoutput=1
\documentclass[11pt]{article}

\usepackage[]{acl}

\usepackage{times}
\usepackage{latexsym}

\usepackage[T1]{fontenc} 
\usepackage[utf8]{inputenc}

\usepackage{microtype}
\usepackage{graphicx}
\usepackage{makecell}
\usepackage{amssymb}
\usepackage{xcolor}
\usepackage{soul}
\usepackage{float}
\usepackage{booktabs}

\newcommand{\minus}{\scalebox{0.4}[1.0]{$-$}}
\newcommand{\nop}[1]{}
\newcommand{\add}[1]{\textcolor{black}{#1}} % red
\newcommand{\mww}[1]{\textcolor{black}{#1}} % purple
\newcommand{\lm}[1]{\textcolor{black}{#1}} %cyan
\newcommand{\mo}[1]{\textcolor{black}{#1}} % purple
\newcommand{\al}[1]{\textcolor{black}{#1}} % pink
\newcommand{\mlb}[1]{\textcolor{black}{#1}}

\title{Towards Transparent Interactive Semantic Parsing\\ via Step-by-Step Correction}

\author{Lingbo Mo, Ashley Lewis, Huan Sun, Michael White \\
        The Ohio State University \\ \texttt{\big\{mo.169,lewis.2799,sun.397,white.1240\big\}@osu.edu }}

\begin{document}
\maketitle

\begin{abstract}

Existing studies on semantic parsing focus on mapping a natural-language utterance to a logical form (LF) in one turn. However, because natural language may contain ambiguity and variability, this is a difficult challenge. In this work, we investigate an interactive semantic parsing framework that explains the predicted LF \textit{step by step} in natural language and enables the user to make corrections through \textit{natural-language feedback} for individual steps. We focus on question answering over knowledge bases (KBQA) as an instantiation of our framework, aiming to increase the transparency of the parsing process and help the user trust the final answer. We construct \textsc{INSPIRED}, a crowdsourced dialogue dataset derived from the \textsc{ComplexWebQuestions} dataset. Our experiments show that this framework has the potential to greatly improve overall parse accuracy. Furthermore, we develop a pipeline for dialogue simulation to evaluate our framework w.r.t.\ a variety of state-of-the-art KBQA models without further crowdsourcing effort. The results demonstrate that our framework promises to be effective across such models.\footnote{Our \textsc{INSPIRED} dataset and code are available at \href{https://github.com/molingbo/INSPIRED}{https://github.com/molingbo/INSPIRED}.}

\end{abstract}
\section{Introduction}
\label{intro}

Semantic parsing aims to map natural language to formal meaning representations, such as $\lambda$-DCS, API calls, SQL and SPARQL queries. As seen in previous work~\cite{liang2013learning, yih2014semantic, yih2015semantic, talmor2018web, chen2019uhop, lan2020query, gu2021beyond}, parsers still face major challenges: (1) the accuracy of state-of-the-art parsers is not high enough for real use, given that natural language questions can be ambiguous or highly variable with many possible paraphrases, and (2) it is hard for users to understand the parsing process and validate the results.

\begin{figure}[!t]
    \centering
    \includegraphics[width=0.85\linewidth]{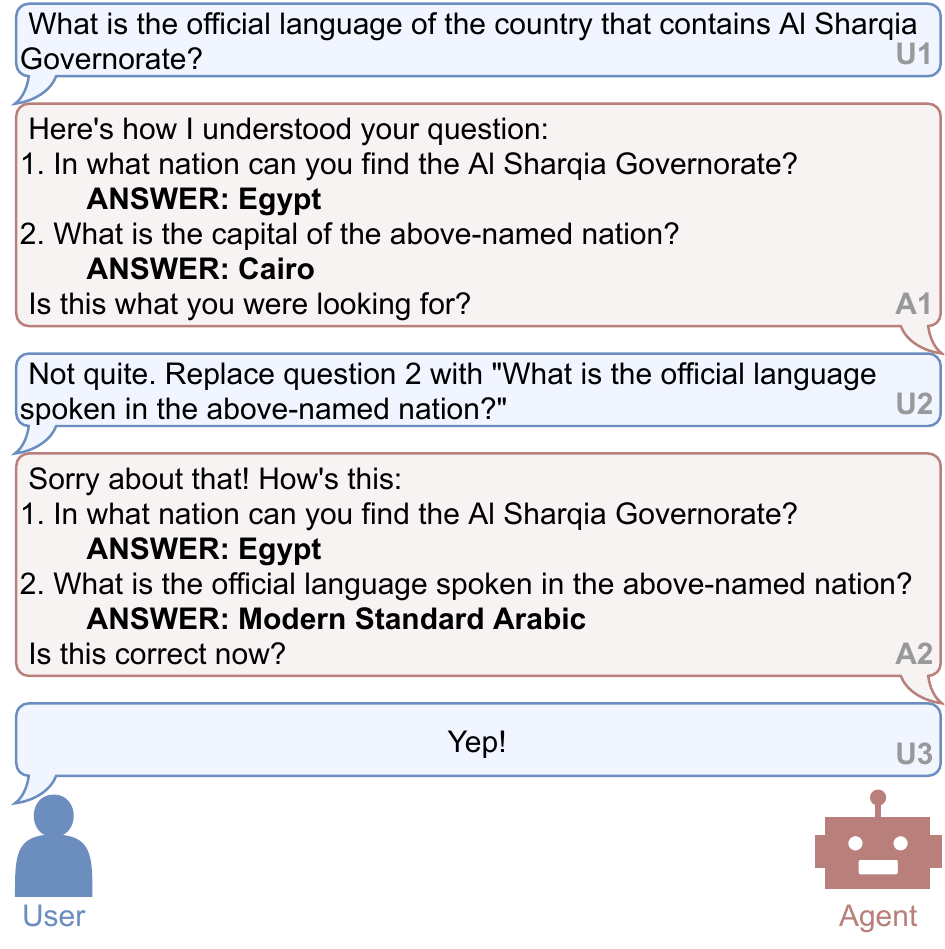}
    \vspace{-10pt}
    \caption{Example dialogue from our dataset (dubbed \textsc{INSPIRED})\nop{the \textsc{INSPIRED} dataset}. The agent turns (A$_i$'s) illustrate our emphasis on transparency by explaining the predicted logical form step by step in natural language, along with intermediate answers, to the user for feedback.}
    \label{fig:dialog_example_1}
    \vspace{-5pt}
\end{figure}

\begin{figure*}[!t]
    \centering
    
    \includegraphics[width=1\linewidth]{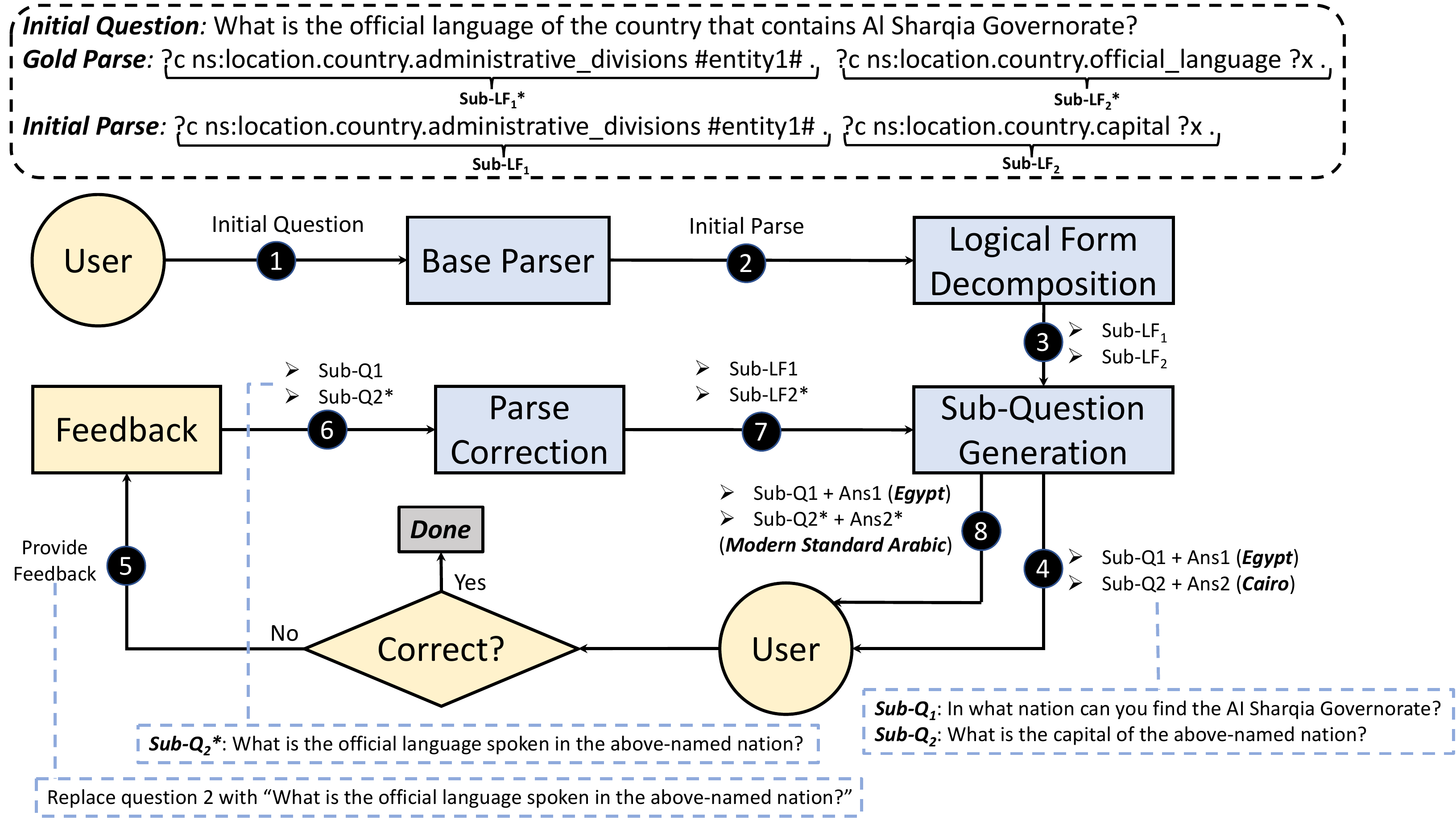}
    \vspace{-5pt}
    \caption{Illustration of our interactive semantic parsing framework for KBQA. The box on the top lists a running example. The prefix of a SPARQL query (i.e., LF used for KBQA in this paper) in this example is omitted for brevity. The figure on the bottom shows the entire workflow of our framework using the example above.}
    \label{fig:full_process}
\end{figure*}

In response to the challenges above, recent work~\cite{li2014constructing, he2016human, chaurasia2017dialog, su2018natural, gur2018dialsql, yao2019interactive, elgohary2020speak} explores \textsl{interactive semantic parsing}, involving human users to give feedback and boost system accuracy. For example, \citet{su2018natural} show that fine-grained user interaction greatly improves the usability of natural language interfaces to Web APIs. \citet{yao2019interactive} allow their semantic parser to ask users clarification questions when generating an If-Then program. And recently, \citet{elgohary2020speak} crowdsources the SPLASH dataset for correcting SQL queries using natural language feedback.

Compared with these approaches, we aim to enhance the transparency of the parsing process and increase user confidence in the final answer. Figure \ref{fig:dialog_example_1} shows a desired dialogue between user and agent. We design an interactive framework for semantic parse correction that can explain the predicted complex logical form (LF) in a step-by-step manner and enable the user to make corrections to individual steps in natural language. To demonstrate the advantages of our interactive framework, we propose to instantiate it for complex question answering over knowledge bases (KBQA), where interactive semantic parsing has remained largely unexplored.

Figure~\ref{fig:full_process} illustrates our framework with a concrete example: A base parser predicts an initial parse, which we decompose into sub-LFs and translate to natural-language questions (i.e., \textit{Sub-Question Generation}). This shows the steps of answering the question, allowing the user to see exactly how a final answer is found and be confident that it is correct or give feedback in natural language to correct the steps. If any user feedback is given, our framework uses it to correct errors in the current parse (i.e., \textit{Parse Correction}).

To build models for Sub-Question Generation and Parse Correction, we construct a dataset via crowdsourcing, based on the \textsc{ComplexWebQuestions} (CWQ) dataset~\cite{talmor2018web}, which is widely used for complex QA. To make LFs understandable to crowdworkers, we translate each sub-LF into a templated sub-question using a rule-based method. During crowdsourcing, workers paraphrase the templated question into a natural one. We create a dialogue for each complex question, an example of which is shown in Figure~\ref{fig:dialog_example_1}. Our dataset, dubbed \textsc{INSPIRED} (\textbf{IN}teractive \textbf{S}emantic \textbf{P}ars\textbf{I}ng for Cor\textbf{RE}ction with \textbf{D}ecomposition), will facilitate further exploration of interactive semantic parsing for KBQA.

Our main contributions are as follows: (1) We design a more transparent interactive semantic parsing framework that explains to a user how a complex question is answered step by step and enables them to make corrections in natural language and trust the final answer. (2) To support research on interactive semantic parsing for KBQA, we release a high-quality dialogue dataset using our framework. (3) We establish baseline models for two core sub-tasks in this framework: Sub-Question Generation and Parse Correction. (4) Although \textsc{INSPIRED} is constructed using a selected base parser, we are able to train models to simulate user feedback, allowing us to study the promise of our framework to correct errors made by other semantic parsers without more annotation effort. With these contributions, we hope to inspire many directions of future work, which we discuss in the end.

\section{Dataset Construction}
\label{dataset_construction}
In this section, we describe the workflow for dataset construction following the design of our framework (Figure~\ref{fig:full_process}). We prepare pairs of complex questions and SPARQL parses predicted by a base semantic parser (Section~\ref{prepare_question}). Then, we decompose the gold and predicted parses and determine \textsl{correction operations} (Section~\ref{query_decomposition}). \lm{The sub-LFs are translated to questions using templates (Section~\ref{explain_sparql}) and we employ crowdworkers to paraphrase these questions using natural language (Section~\ref{crowdsourcing}).}

\subsection{Dialogue Preparation for Crowdsourcing}

\label{dialogue_preparation}
\subsubsection{Preparing Questions and SPARQL}
\label{prepare_question}

We start with the \textsc{ComplexWebQuestions 1.1} (CWQ) dataset \cite{talmor2018repartitioning, talmor2018web}, which contains complex questions paired with gold SPARQL queries for Freebase~\cite{bollacker2008freebase}. We adopt a transformer-based seq2seq model~\cite{vaswani2017attention} as the base semantic parser to prepare a predicted SPARQL query for each complex question (see the second and third paragraphs in Section~\ref{sec:experiments} for our rationale).

As a simplifying assumption, we take gold named entities mentioned in a question as given. Specifically, we replace named entities in a SPARQL query with special tokens such as \textsl{\#entityX\#}, where \textsl{X} is a number corresponding to the order in which the entity appears. After parsing, we replace these tokens with the gold entities. The challenge of addressing errors caused by named entity recognition and linking in a real KBQA system is left as an important piece of future work.

In order to reduce data collection cost, we select a subset of questions in the training data of CWQ to create dialogues in \textsc{INSPIRED}'s training set. We conduct an analysis of repeated predicates and question types, and ensure that each predicate occurs at least three times in \textsc{INSPIRED}'s training set when possible.  We include every question where the base parser makes an error and ensure coverage of the four multi-hop reasoning types~\cite{talmor2018web}. Different reasoning types require different translation strategies in order to represent their logical forms in English (see Section~\ref{explain_sparql} and Appendix~\ref{DC_types}). We create 10,374 dialogues in total, based on 3,492 questions from the training set, 3,441 from the validation set, and 3,441 from the test set of CWQ. We omit a small set of questions from the original validation and test sets that are consistently confusing to crowdworkers. Table \ref{tab:stats} shows a breakdown of the CWQ question types in the INSPIRED dataset, along with the average number of corrections and sub-questions.

\begin{table}
\begin{center}
\resizebox{0.9\linewidth}{!}{
\begin{tabular}{l c c c | c}
\toprule

\textbf{Number of} & \textbf{Train} & \textbf{Dev} & \textbf{Test} & \textbf{Overall} \\
\midrule
Complex Questions &  3,492 & 3,441 & 3,441 & 10,374 \\
- Composition & 1,196 & 1,532 & 1,490 & 4,218 \\
- Conjunction & 1,796 & 1,503 & 1,553 & 4,852 \\
- Comparative & 253 & 217 & 207 & 677\\
- Superlative & 247 & 189 & 191 & 627\\
\al{Predicted Sub-Questions} & 1.7 & 2.0 & 1.9 & 1.9\\
\al{Gold Sub-Questions} & 2.2 & 2.1 & 2.1 & 2.1\\
\hline
\multicolumn{3}{l}{\al{Range of the number of predicted sub-questions}} &  & 0 - 5 \\
\multicolumn{3}{l}{\al{Range of the number of gold sub-questions}} &  & 2 - 4 \\
\multicolumn{3}{l}{Average number of edits} &  & 1.4 \\
\multicolumn{3}{l}{Dialogues with 0 edits} &  & 5,016 \\

\bottomrule
\end{tabular}}
%\vspace{-5pt}
\caption{Statistics for our \textsc{INSPIRED} dataset: the number of complex questions for each reasoning type, the average number of sub-questions and edit operations in a dialogue (excluding those that do not have edits).} \label{tab:stats}
\end{center}
%\vspace{-15pt}
\end{table}

\subsubsection{Logical Form Decomposition}
\label{query_decomposition}

An important goal of creating \textsc{INSPIRED} is to make the process of question answering transparent to the user. Each dialogue features a decomposition process by which our framework transforms the complex question into an initial parse, breaks it into sub-LFs, retrieves answers, and presents this whole process to the user for correction. The overarching strategy of the decomposition process is to identify the predicates that express distinct components in the LF of the complex question, which correspond to individual sub-questions. Typically, these components appear as a \textsl{triple} in the logical form such as \textsl{Sub-LF$_{1}$} in Figure~\ref{fig:full_process}, which is comprised of a head entity, a predicate, and a tail entity. Logical forms in CWQ typically contain two or three of these components. There can be multiple predicates that group together to express one component, for example those connected by a CVT\footnote{CVT is a Type within Freebase which is used to represent data where each entry consists of multiple fields.} (Compound Value Type) node, in which case the two predicates and their two entities will form one component. Within these, there can be filters and/or restrictions, which provide additional information about entities of the main predicate and are typically merged to the corresponding component. Details about how we deal with these logical form components are in Appendix \ref{DC_LF_features} and more concrete examples about decomposition are shown in Table~\ref{tab:question_type}.

Using this strategy, we decompose both the parser’s predicted SPARQL query and the gold one into sub-LFs, and compare those sub-LFs to determine the sequence of operations needed to transform the predicted parse into the gold parse, including inserting, deleting, or replacing a sub-LF, which is to be paraphrased by crowdworkers (Section~\ref{crowdsourcing}) based on our templated sub-question. These operations determine the ``correction'' steps in each dialogue, where the agent asks the user if any corrections are needed (Figure \ref{fig:dialog_example_1}, turn A1), and the user either confirms that the initial parse is correct or provides corrections (turn U2). Though any new sub-questions that are introduced use natural and varied language, the correction operations are given using templates (i.e., \textit{replace question \#X with Y, delete question \#X, insert question Y}). More details about how dialogues are formed around complex questions can be found in Appendix \ref{AP1_selection_decomposition}.

\subsubsection{Explaining SPARQL}
\label{explain_sparql}

We develop a strategy for how to represent the sub-SPARQL queries in a form that crowdworkers can understand after decomposition, for which we create a template corpus and a rule-based translation method to do so. The corpus consists of 772 different predicates that appear in the CWQ dataset and translations of each into a basic template that conveys the content. More details about the translation of LFs with different reasoning types into sub-questions are found in Appendix \ref{DC_translation} and \ref{DC_types}.

\subsection{Crowdsourcing}
\label{crowdsourcing}

To make queries understandable for an average user, as in Figure \ref{fig:dialog_example_1}, we translate the decomposed LFs into English questions using templates as mentioned in Section~\ref{explain_sparql}. To obtain natural sounding questions, we conduct crowdsourcing on Amazon Mechanical Turk (AMT), in which crowdworkers are employed to rephrase sub-questions from the clunky, templated form into more concise and natural English in the context of a dialogue. The task is conducted using ParlAI~\cite{miller2017parlai}, which allows us to set up a versatile dialogue interface.

In each dialogue, every turn of the interlocutors has prescribed content. A total of 14 crowdworkers are employed to express the content in natural language and complete a maximum of 1,800 dialogues. Because the crowdsourcing task for this dataset requires extensive, detailed instructions, we design the task quite carefully with multiple stages of checkpoints to ensure quality of data collection. An overview of these phases can be seen in Table~\ref{tab:crowdsourcing_protocol} and other details are presented in Appendix~\ref{AP1_crowdsource}. We recruit and retain a small set of exemplary workers for this task (see item 4 in General Principles in Table~\ref{tab:crowdsourcing_protocol}). This phased strategy, while requiring more effort, proves to be effective in ensuring overall data quality which will be shown in Section~\ref{dataset_analysis}. 

\begin{table}[t!]
\large
\begin{center}
\resizebox{1\linewidth}{!}{
\begin{tabular}{|l|}

\hline

\makecell{\textbf{Phased Crowdsourcing Protocol}} \\ 
 
\textbf{Phase 1: Tutorial}\\
\makecell[l]{1. Worker reads examples and explanations of the task.}\\
\makecell[l]{2. Worker receives specific instructions for how to rephrase questions of \\ different types.}\\

\textbf{Phase II: Qualification Quiz}\\
\makecell[l]{1. Worker completes an 8-question multiple choice quiz. Quiz questions \\ are based on the tutorial content.}\\
\makecell[l]{2. Worker must achieve at least 7 out of 8 to pass. They may take the quiz \\ more than once, but there is a ten minute wait period between attempts.}\\

\textbf{Phase III: Trial Period}\\
\makecell[l]{1. Worker completes 10 predetermined tasks which were chosen as \\ representative examples for all the tasks.}\\
\makecell[l]{2. Tasks are manually graded. If the work is overall good, the worker receives \\ specific feedback on anything that was done incorrectly.}\\
\makecell[l]{3. If quality is not good, worker is eliminated.}\\
\makecell[l]{4. Workers get paid the regular rate for each task and upon completing the 10 \\ tasks, receive a bonus for the time spent on the tutorial and qualification quiz.}\\

\textbf{Phase IV: Batches of Tasks}\\
\makecell[l]{1. Worker is given access to a batch of 100 tasks, which are spot-checked for \\ quality. A bonus is given as the worker passes each set of 100 tasks.}\\
\makecell[l]{2. If quality is good, workers are given a second batch of 100 questions, also \\ spot checked.}\\
\makecell[l]{3. Batch size increases based on worker quality and speed.}\\
\makecell[l]{4. Worker completes up to 1800 tasks.}\\
 
\textbf{General Principles}\\
\makecell[l]{1. Prompt feedback, payment, and release of new batches
}\\
\makecell[l]{2. Provide a link to the tutorial so that it can be accessed at any time.}\\
\makecell[l]{3. Higher than average payment.}\\
\makecell[l]{4. Keep pool of workers small for better communication and quality control.}\\
\makecell[l]{5. Verify that workers are native English speakers.}\\
 
\hline

\end{tabular}}
\vspace{-5pt}
\caption{The phased crowdsourcing protocol for our Amazon Mechanical Turk task.}

\label{tab:crowdsourcing_protocol}
\end{center}
\vspace{-15pt}
\end{table}

\section{Dataset Analysis}
\label{dataset_analysis}

In this section, we conduct a thorough quality analysis of \textsc{INSPIRED} dataset and highlight aspects that contribute to overall quality, including paraphrasing characteristics and contextual awareness.

\textbf{Overall Data Quality.} In each dialogue, the crowdworker is required to rephrase the original complex question and each templated sub-question. Overall, we believe the quality of the data to be high for a few reasons. In the collection process, our crowdworkers read a detailed tutorial, pass two qualification tasks, and have their work spot-checked at each stage of collection. Because we keep our pool of workers small, we are able to maintain frequent communication with them throughout the process, giving feedback in an ongoing fashion. 

Furthermore, we use a semi-automatic data cleaning method to identify inaccurate paraphrases for manual repair, resulting in edits to 325 sub-questions in total. Based on our observation on a held-out subset of the data, we estimate that only 3.1\% of all sub-questions still have inaccuracies, after cleaning. More details are in Appendix \ref{AP1_data_cleaning}.

\begin{table}[t!]
\begin{center}
\resizebox{1\linewidth}{!}{
\begin{tabular}{ccc}
\toprule

 & \textbf{Template Corpus} & \textbf{Rephrased Corpus} \\
\textbf{Avg Length} & 17.3 & 10.7 \\
\textbf{Unigrams} & 8,465 & 9,864 \\
\textbf{Bigrams} & 21,072 & 44,085 \\
\textbf{Trigrams} & 31,838 & 81,479 \\

\bottomrule
\end{tabular}
}
\vspace{-5pt}
\caption{Comparison of average length (in words) of templated and rephrased questions as well as the size of vocabulary for 1-, 2-, and 3-grams across all templates and rephrased questions, demonstrating the increased diversity of rephrased questions.}

\label{tab:diversity_metrics}
\end{center}
\vspace{-15pt}
\end{table}

\textbf{Paraphrasing Characteristics.}
Table \ref{tab:diversity_metrics} shows the difference between the vocabularies (unique words) of all the templates in \textsc{INSPIRED} and the rephrased versions of sub-questions, which are calculated using GEM evaluation scripts \cite{gehrmann2021gem}. Further, the mean length of the templated questions is 17.3 words, while the mean length of the rephrased questions is 10.7 words. These comparisons demonstrate that the rephrased questions show much more diversity in phrasings and lexical choices, but are also more concise. More GEM metrics can be seen in Appendix \ref{sec:appendix2}. 

In order to better understand how crowdworkers rephrased templates, 100 randomly selected sub-questions are studied in terms of lexical relationships between the template and rephrased versions. We find that they are using synonmy, hypernymy and hyponymy in rephrasings of the templates, in addition to changing word order. This analysis can be found in Appendix \ref{sec:lexical}.

\textbf{Contextual Awareness.} Additionally, crowdworkers are encouraged to incorporate contextual information of a given sub-question into their rephrasings, thus improving the contextual richness of the dataset. In order to demonstrate contextual awareness, Table \ref{tab:context_aware} shows the average ROUGE-1 and ROUGE-2 scores of all sub-questions in their actual contexts (the complex question and any preceding sub-questions), in comparison to the same sub-questions in a randomly assigned context that utilizes the same sub-logical form. Entities are masked with \textsl{\#entity\#} tokens to prevent the actual context from being advantaged by overlap in entity names. The higher scores for the actual context indicate that the wording of sub-questions reflect the context from which they are derived.

\mlb{In general, it is natural for human users to consider the context when making utterances in a dialogue. From the perspective of model development, providing contextual information enriches the input by providing relevant information that may not be present in a given sub-question or sub-LF. We provide concrete examples and analysis to show the effect of context dependency in Table~\ref{tab:ROUGE_examples} in Appendix \ref{contextual_awareness}. Moreover, experiments considering different contexts in Section~\ref{sec:experiments} further validate the impact of context dependence on parse correction and sub-question generation performance.}

\begin{table}[t!]
\begin{center}
\resizebox{0.8\linewidth}{!}{
\begin{tabular}{l|cc}
\toprule
\textbf{} & \textbf{ROUGE-1 } & \textbf{ROUGE-2 } \\
\midrule
Random Context & 22.8 & 3.4 \\
Actual Context & \textbf{27.7} & \textbf{6.2} \\
\bottomrule
\end{tabular}
}
\vspace{-5pt}
\caption{Comparison of the n-gram overlap between the paraphrase and the context for a sub-LF vs.\ other randomly chosen context for the same sub-LF.}
\label{tab:context_aware}
\end{center}
\vspace{-15pt}
\end{table}

\section{Experiments}
\label{sec:experiments}

In this section, we explore several base semantic parsers and show how we choose one as the initial parser to construct \textsc{INSPIRED}. Then, we conduct extensive experiments on those two core sub-tasks (i.e., sub-question generation and parse correction) in our framework. Finally, in order to study the promise of our framework for other parsers (beyond the one used to construct \textsc{INSPIRED}) without introducing extra crowdsourcing effort, we simulate dialogues based on our trained models for sub-question generation and parse correction. We train all models on 4 GTX 1080 Ti 11 GB GPUs.

Firstly, we explore Transformer~\cite{vaswani2017attention}, BART-large~\cite{lewis2020bart} and QGG~\cite{lan:acl2020} as base parsers. In the official leaderboard\footnote{\url{https://www.tau-nlp.org/compwebq-leaderboard}} of CWQ, QGG is the best-performing method in the line of query graph generation approaches. Models like NSM+h~\cite{he2021improving} and PullNet~\cite{sun2019pullnet} directly output final answers without LFs, which cannot be made more transparent or interactive with our framework. CBR-KBQA~\cite{das2021case} is the SOTA model on this dataset \add{as of the submission time}, but as its code is not available, we choose Transformer and BART-large as the two candidate parsers. We input the complex question to these two seq2seq models and output the LF. Since entities are masked in the LFs for these models, we provide QGG with gold entities for fair comparison. We report their LF exact match (EM) and F1 scores in Table~\ref{tab:parse}.

\begin{table}[t!]
\begin{center}
\resizebox{0.9\linewidth}{!}{
\begin{tabular}{l|cc}
\toprule
\textbf{Models} & \textbf{EM} & \textbf{F1} \\
\midrule
*Transformer \cite{vaswani2017attention}  & 52.3 & 58.6 \\
BART-large \cite{lewis2020bart} & 60.9 & 65.8 \\
QGG \cite{lan:acl2020} & - & 49.0 \\ 
\bottomrule
\end{tabular}}
\vspace{-5pt}
\caption{\mo{Performance of different semantic parsers} on CWQ test set.\protect\footnotemark\ The asterisk (*) denotes the initial semantic parser we choose \mo{for constructing \textsc{INSPIRED}.}}

\label{tab:parse}
\end{center}
\vspace{-15pt}
\end{table}

\footnotetext{Since \textsc{INSPIRED} excluded a small set of questions from CWQ, for fair comparison, scores here are calculated using questions in CWQ test set which are included in \textsc{INSPIRED}.}

We finally select Transformer as the initial parser because it is neither state-of-the-art nor has overly poor performance. As the intention is to create a dataset that represents a wide range of parsing errors and correction strategies, a ``middle-of-the-road'' parser is best for achieving good coverage but also being of decent quality. We report the characteristics of errors made by Transformer in Appendix ~\ref{error_char}. We will explore the other two models in Table~\ref{tab:parse} through simulation (Section \ref{simulation}).

In the following two sections, we explore two sub-tasks under our framework. We treat both of them as seq2seq tasks, then present and evaluate several baseline models including Seq2Seq~\cite{sutskever2014sequence}, Transformer~\cite{vaswani2017attention}, BART-base and BART-large~\cite{lewis2020bart} for each task, in which we use \textsc{INSPIRED} for training and testing. After that, we conduct error analysis for both sub-tasks by examining 100 examples respectively. Details of the analysis are in Appendix \ref{sec:appendix4}.

\subsection{Parse Correction with NL Feedback}

Given a sub-question $q$, the parse correction task is to convert it into a new sub-LF $p$. By parsing the templates used by correction operations as mentioned in Section~\ref{query_decomposition}, we extract the operation (i.e., replace, delete, or insert a sub-question) and apply it to the appropriate step. Then, sub-LFs are compiled accordingly to form a correction parse $P$ for the entire question. We predict the sub-LF based on $q$ without considering contexts, and present the results of several baselines in Table~\ref{tab:parse_1_1}. We report both the turn-level accuracy---the accuracy of sub-LFs in correction turns---and the dialog-level accuracy---the end-to-end accuracy of the entire LFs after correction---on our test set. 

Since models like BART adopt a subword tokenization scheme, the validness of predicates generated by concatenating subwords can not always be guaranteed. We use beam search of size 10 to generate LFs as candidates, filtering those with invalid predicates and excluding erroneous predictions previously made by the parser. We additionally compare with a baseline named 2nd-Beam, which applies beam search on the base parser to obtain two initial parses and uses the second for parse correction. It has some performance gains over the setting without correction, but is much lower than those settings with human feedback. Results in Table~\ref{tab:parse_1_1} further suggest: (1) incorporating human feedback can substantially improve the parse accuracy and (2) using BART-large with pretraining as the correction model achieves the best performance, achieving 19.0 points higher than the initial parser in terms of the dialog-level EM score.

\begin{table}[t!]
\begin{center}
\resizebox{1\linewidth}{!}{
\begin{tabular}{l|cc}
\toprule
\textbf{Correction Models} & \textbf{ \makecell[c]{Turn-level EM } } & \textbf{ \makecell[c]{Dialog-level EM} } \\
\midrule
w/o Correction & - & 52.3 \\
2nd-Beam & - &  55.8 \\
\midrule
Seq2Seq(LSTM) & 78.9 & 65.0 \\
Transformer & 81.2 & 68.0 \\
BART-base & 82.3 & 70.3 \\
BART-large & \textbf{82.9} & \textbf{71.3} \\
\bottomrule
\end{tabular}
}
\vspace{-5pt}
\caption{Turn-level and Dialogue-level accuracy of different models after incorporating feedback (where applicable).}
\label{tab:parse_1_1}
\end{center}
\vspace{-15pt}
\end{table}

Then, using BART-large as the correction model, we further study the correction process by concatenating different contexts to the input, including the history of sub-questions $h_{q}$ and sub-LFs $h_{\mathit{lf}}$. We report both the accuracy for each turn of correction and the end-to-end accuracy. As shown in Table~\ref{tab:parse_1_2}, we find that: (1) Adding contexts into the input can further improve the correction accuracy. (2) As the number of turns goes up, context contributes more to the correction process, which indicates that including the full dialogue history in the input leads to the best results. (3) The BART-large model with inputs that leverage $h_{q}$ and $h_{\mathit{lf}}$ achieves the best performance, with a 21.2 increase under dialog-level EM compared to the initial parser.

\begin{table}[t!]
\begin{center}
\resizebox{1\linewidth}{!}{
\begin{tabular}{l|c|cccc}
\toprule
\textbf{Context} & \textbf{ \makecell[c]{Dialog-level \\ EM} } & \textbf{ \makecell[c]{Turn-1\\(3441)} } & \textbf{ \makecell[c]{Turn-2\\(3441)} }  & \textbf{ \makecell[c]{Turn-3\\(345)} } & \textbf{ \makecell[c]{Turn-4\\(56)} } \\
\midrule
w/o Correction & 52.3 & - & - & - & - \\
\midrule
\textbf{BART-large} &  &  &  &  &  \\
w/o Context & 71.3 & 84.6 & 81.5 & 85.5 & 53.6 \\
+ $h_{\mathit{q}}$ & 72.2 & 84.7 & 82.2 & 89.3 & \textbf{100.0} \\
+ $h_{\mathit{lf}}$ & 72.0 & 84.3 & 82.1 & 89.3 & \textbf{100.0}  \\
+ $h_{\mathit{q}}$ \& $h_{\mathit{lf}}$ & \textbf{73.5} & \textbf{86.4} & \textbf{83.2} & \textbf{91.0} & \textbf{100.0}  \\
\bottomrule

\end{tabular}
}
\vspace{-5pt}
\caption{Parse correction performance when considering different contexts. $h_{\mathit{lf}}$ and $h_{\mathit{q}}$ denote the dialogue history of sub-LFs and sub-questions respectively.}

\label{tab:parse_1_2}
\end{center}
\vspace{-15pt}
\end{table}

\subsection{Sub-Question Generation}
\label{LF2NL}

Sub-question generation aims to translate a sub-LF $p$ into a natural sub-question $q$. Table~\ref{tab:generation_1_1} lists generation performance from five baselines without considering contexts. We explore an off-the-shelf paraphrasing model,\footnote{https://huggingface.co/eugenesiow/bart-paraphrase} which takes corresponding templated sub-question $q^{t}$ as the input and outputs $q$. It is fine-tuned on BART-large using three paraphrasing datasets including Quora,\footnote{https://www.kaggle.com/c/quora-question-pairs}  PAWS~\cite{zhang2019paws} and MSR paraphrase corpus~\cite{dolan2005automatically}. The low scores demonstrate that sub-question generation is more challenging than a simple paraphrasing task. For the other models, we explore two scenarios with different inputs: (1) sub-LF $p$ only and (2) a concatenation of $p$ and $q^{t}$. We report BLEU scores based on n-grams overlap and BERTScores measuring semantic similarity. The results in Table~\ref{tab:generation_1_1} suggest that: (1) Using BART-large as the generation model achieves the best performance and (2) incorporating the templated sub-questions into the model input can improve performance on all baselines, which makes sense because some tokens in $q^{t}$ can be directly copied into the output question.

\begin{table}[t!]
\begin{center}
\resizebox{0.85\linewidth}{!}{
\begin{tabular}{l|ccc}
\toprule
\textbf{Generation Models} & \textbf{BLEU-2} & \textbf{BLEU-4} & \textbf{BERTScore} \\
\midrule
BART-paraphrase & 10.6 & 2.7 & 88.0 \\
\midrule
Seq2Seq(LSTM)  & 17.8  & 6.4  & 90.8  \\
Seq2Seq(LSTM)$^{t}$  & 18.7 & 6.7  & 91.3  \\ \midrule
Transformer & 21.1 & 8.4 & 91.7 \\
Transformer$^{t}$  & 23.4  & 9.1  & 92.6  \\ \midrule
BART-base & 30.7 & 15.0  & 93.8 \\
BART-base$^{t}$  & 32.0 & 15.9  & 94.1  \\ 
\midrule
BART-large & 31.5 & 15.4 & 94.0 \\
BART-large$^{t}$  & \textbf{32.4} & \textbf{16.2} & \textbf{94.2}\\
\bottomrule
\end{tabular}
}
%\vspace{-5pt}
\caption{Question generation performance of different models. $t$ denotes that the input incorporates templated sub-question,  as well as the current sub-logical form.}
\label{tab:generation_1_1}
\end{center}
\vspace{-10pt}
\end{table}

\begin{table}[t!]
\begin{center}
\resizebox{0.85\linewidth}{!}{
\begin{tabular}{l|ccc}
\toprule
\textbf{Context} & \textbf{BLEU-2} & \textbf{BLEU-4} & \textbf{BERTScore} \\
\midrule
\textbf{BART-large$^{t}$} & & &  \\
w/o Context & 32.4 & 16.2 & 94.2  \\

+ $h_{q^{t}}$ & 33.3  & 16.5 & 94.6  \\
+ $Q$ & 33.4 & 16.6 & 94.6  \\

+ $Q$ \& $h_{q^{t}}$  & \textbf{34.1} & \textbf{17.1} & \textbf{94.8}  \\

\bottomrule
\end{tabular}
}
\vspace{-5pt}
\caption{Comparison of question generation performance when considering different contexts in the input.}
\label{tab:generation_1_2}
\end{center}
\vspace{-10pt}
\end{table}

Furthermore, we use the best-performing model (i.e. BART-large with both $p$ and $q^{t}$ as the input) in Table~\ref{tab:generation_1_1} as the basic setting to explore the modeling of different contexts including the complex question $Q$ and the history of templated sub-questions $h_{q^{t}}$. As shown in Table~\ref{tab:generation_1_2}, we find that (1) adding context into the model's input can obtain higher metric scores, which suggests that context can help in a dialogue. (2) Those settings that incorporate the original complex question $Q$ generally perform better than the others, since the complex question contains the semantics of the sub-question to be generated. (3) BART-large with the input containing both $Q$ and the history of templated sub-questions achieves the best performance. We also tried incorporating the history of sub-LFs $h_{\mathit{lf}}$, but it does not help further improve the performance.

Because automatic metrics like BLEU scores do not necessarily paint a full picture of the model performance, we manually check 100 generated questions. They are indeed of high quality and semantically similar to the human-written ones; see details in the second part of Appendix~\ref{sec:appendix4}.

\subsection{Simulation}
\label{simulation}
In this section, we demonstrate that our framework can pair with other KBQA parsers and use simulated user feedback to correct their errors. To simulate a dialogue, we develop a pipeline: (1) Automatically translate a parser's predicted LFs into natural questions using the sub-question generation model equipped with the best-performing setting in Table~\ref{tab:generation_1_2}. (2) Use oracle error detection and train a generator to simulate a human user's corrections for these dialogues. This generator is a BART-large model that leverages the complex question and templated sub-questions as input to generate human feedback. (3) Correct erroneous parses using the previously trained parse correction model under the best-performing setting in Table~\ref{tab:parse_1_2}.

\begin{table}[t!]
\small
\centering
\begin{minipage}{0.45\linewidth}
\resizebox{1\linewidth}{!}{
\begin{tabular}{l|c|c}
\toprule
 & \textbf{BART-large} & \textbf{QGG} \\
\midrule
EM & 60.9 & - \\ 
EM* & \textbf{75.1} & - \\
\midrule
F1 & 65.8 & 49.0 \\
F1* & \textbf{75.7} & \textbf{56.5} \\

\bottomrule
\end{tabular}}
\label{tab:simulation}
\end{minipage}
\hfil
\begin{minipage}{0.45\linewidth}
\resizebox{1\linewidth}{!}{
\begin{tabular}{c|cc}
\toprule
\textbf{Attempt} & \textbf{EM} & \textbf{F1} \\
\midrule
\textbf{BART-large} & &   \\
1   & 75.1 & 75.7 \\
2  & 78.7 & 79.9 \\
3  & \textbf{79.0} & \textbf{80.1} \\ 
\bottomrule
\end{tabular}}
\end{minipage}

\caption{The left table shows the performance of two types of semantic parsers after correction through simulation process, BART-large and QGG. * denotes results after correction. The right table shows BART-large's performance after multiple attempts of correction.}
\vspace{-15pt}
\label{tab:simulation_merge}
\end{table}

We conduct simulation experiments on BART-large~\cite{lewis2020bart} and QGG~\cite{lan:acl2020} respectively from two mainstream methodologies for KBQA as mentioned. We report both F1 and EM for BART-large before and after the correction process using the simulation pipeline. For QGG, since its generated query graphs do not take exactly the same format as SPARQL queries, we report F1 score only. As shown in the left part of  Table~\ref{tab:simulation_merge}, BART-large achieves a 14.2 EM and 9.9 F1 score gain after correction. Meanwhile, the correction process brings 7.5 F1 score improvement for QGG. The results show that \textsc{INSPIRED} can help train effective sub-question generation and parse correction models, which makes our framework applicable to KBQA parsers beyond the one used for constructing \textsc{INSPIRED}. Simulating user feedback makes it easy and far less costly to understand the potential of any base parser (as long as it outputs LFs) under our framework.

Moreover, we expand the simulation experiment to include multiple attempts of correction to simulate situations in which the model does not repair the parse correctly on the first attempt. We use the same human feedback generator to decode several of the highest scoring sequences as candidates for different attempts at correction. We evaluate this strategy after a maximum of three attempts.

Given that sequences decoded by plain beam search~\cite{sutskever2014sequence} often differ only slightly from each other, we adopt diverse beam search~\cite{vijayakumar2018diverse} instead to decode more diverse feedback. As shown in the right part of Table~\ref{tab:simulation_merge}, F1 scores are up to 80.1 after three attempts of correction. We expect CBR-KBQA (the SOTA model mentioned earlier) to do even better given the advantages it has over plain seq2seq models. For example, their retrieval module can alleviate errors caused by sparse predicates. We envision the combination of our framework and theirs as interesting future work.

\section{Related Work}

\noindent \textbf{Conversational Semantic Parsing.}
Conversational semantic parsing (CSP) is the task of converting a sequence of natural language utterances into LFs through conversational interactions. It has been studied in task-oriented dialogues, question answering and text-to-SQL. In task-oriented systems, datasets like MWoZ~\cite{budzianowski2018multiwoz, eric2020multiwoz} and SMCalFlow~\cite{andreas2020task} help users with a specific task (e.g., booking a hotel). CSQA~\cite{saha2018complex} and CoQA~\cite{reddy2019coqa} are built for conversational systems to answer inter-related, simple questions. Meanwhile, ATIS~\cite{hemphill1990atis, dahl-etal-1994-expanding}, SPARC~\cite{yu2019sparc} and CoSQL~\cite{yu2020cosql} are constructed for conversational text-to-SQL tasks. Our work shares a similar objective, i.e., how to represent natural language utterances while considering the multi-turn dynamics of the dialogue. We differ from them in that our task aims at soliciting and applying human feedback to correct generated initial parses.

\noindent \textbf{Interactive Semantic Parsing.}
Multiple works have studied involving human feedback in the parsing process.  \citet{gur2018dialsql} ask multiple choice questions about a limited set of predefined errors. \citet{yao2019model} ask yes/no questions about the presence of SQL components when generating one component at a time. \citet{elgohary2020speak} introduce \textsc{SPLASH}, a dataset for correcting parses in text-to-SQL with free-form natural language feedback. They observe that most mistakes made by neural text-to-SQL parsers are minor, which correspond to editing a schema item (table or column name), a SQL keyword, etc. They can thus be resolved by simply editing a single token or two. Corrections in \textsc{SPLASH} are given in one turn and applied to the entire initial parse. \citet{elgohary-etal-2021-nl} convert feedback in SPLASH into a canonical form of edits that are deterministically applied.

In contrast, we find that parse errors in KBQA are more challenging to resolve. KB relations like `location.country.capital' need to be correctly identified among thousands of candidates, while the table schema in \citet{elgohary2020speak} usually contains only a few table/column names. To make error correction easier in this setting, we break down the parse into a sequence of sub-components and enable the user to provide step-by-step feedback, thereby simplifying the task of parse correction and increasing the likelihood of an accurate parse.

\noindent \textbf{Question Decomposition.}
Question decomposition has been successfully used in complex QA. \citet{iyyer2016answering} propose to answer questions based on tables by decomposing them into inter-related simple questions. \citet{talmor2018web} and \citet{min2019multi} train a model directly to produce sub-questions using question spans. Recent works~\cite{wang2020r3,wolfson2020break} introduce explicit annotation for the decomposition of multi-hop questions into a series of atomic operations.
\citet{wolfson2020break} construct the BREAK dataset and propose QDMR, where questions are decomposed into a series of simpler atomic textual steps. QDMR is an intermediate representation of natural language and LFs, and is not executable on knowledge bases. In our work, we decompose the LF of the complex question into sub-components, which can be directly executed on the KB to retrieve answers. Moreover, we use decomposition to correct the initial parse at a finer-grained level.

\section{Discussion and Future Work}
\label{user_study}

\begin{figure*}[!t]
    \centering
    \includegraphics[width=0.8\linewidth]{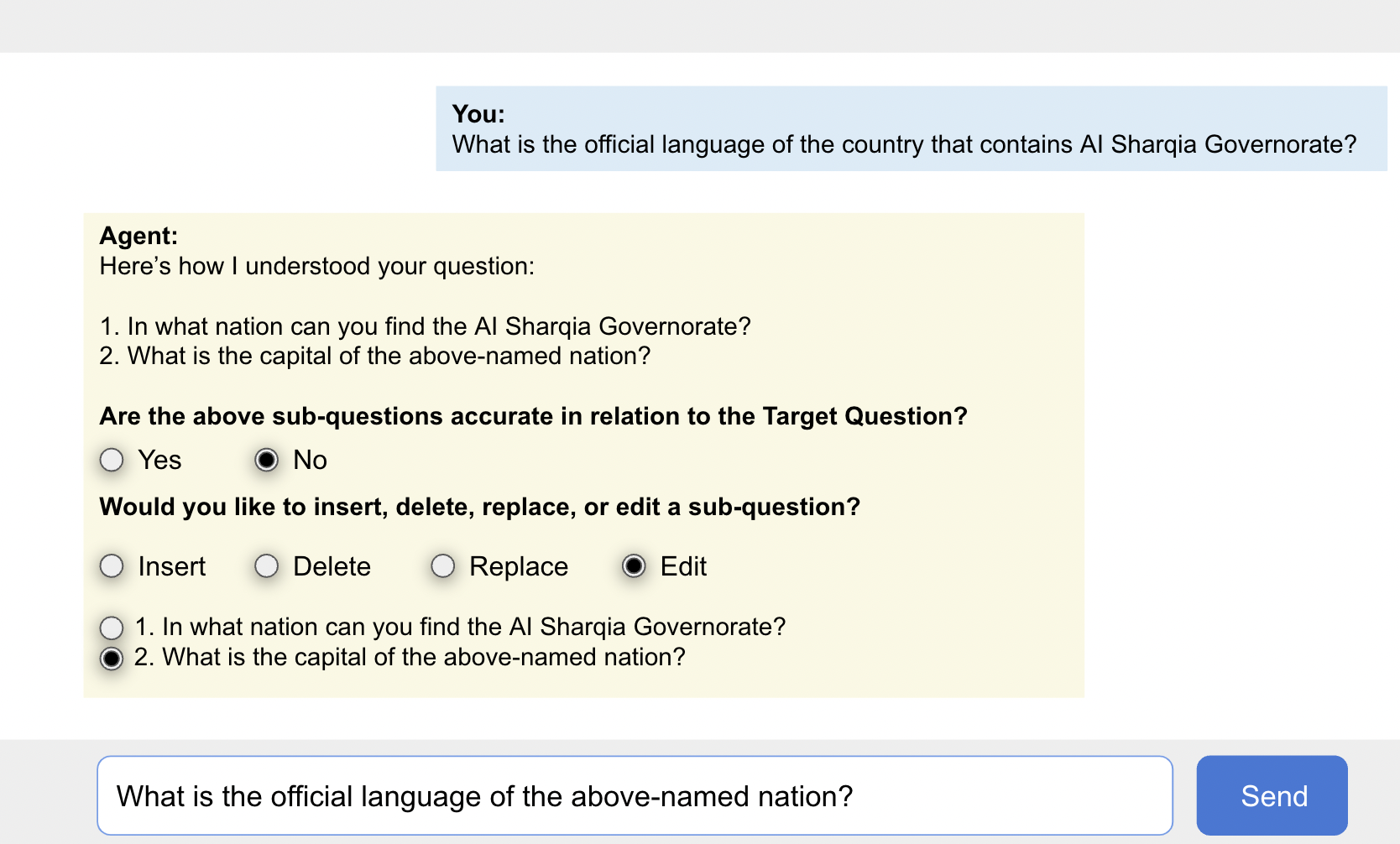}

    \caption{User study interface. \mlb{In addition to inserting/deleting/replacing sub-questions, we provide a new operation `edit' to support minor changes, where the original sub-question is auto-filled into the response box after the user makes the selection. In this example, the user only needs to change \textit{capital} into \textit{official language}.}}
    \label{fig:user_study}

\end{figure*}

\mlb{We are planning to conduct a user study to validate our framework's viability for real use. In this study, human users will utilize the framework to correct parsing errors and query a knowledge base for answers in real time. As shown in Figure~\ref{fig:user_study}, users can specify edit operations in a couple clicks, then type in the response box to insert, replace or edit a sub-question. Note that we add a new `edit' operation to make it easier for users to enter replacement sub-questions that require only small edits.
}

\mlb{While we acknowledge that in a spoken dialogue system, pure natural language feedback may be the most natural, such a system may also suffer from errors caused by automatic speech recognition (ASR)~\cite{wang2020asr, chang21_interspeech}. By contrast, our interface design allows the user to partially specify feedback operations through mouse clicks, which can help mitigate this issue. To evaluate the system, we will use parse accuracy after correction to verify the usefulness of human feedback. We will also use survey questions to measure the subjective quality of the generated explanations, intermediate and final answers, accessibility of the system, etc.}

In this work, the \textsc{INSPIRED} dataset and experiments provide a foundation for many directions of future work. For example, this could take the shape of gains in parse accuracy as well as improvements to the correction strategy through decomposition. The simulation pipeline provided can also be used for further experimentation. Other complementary work could include handling errors introduced by named entity recognition and linking. Lastly, applying our framework to other query languages like SQL could be an exciting direction.

\section{Conclusion}

We have proposed an interactive semantic parsing framework and instantiated it with KBQA in this work.  Using this framework, we crowdsourced a novel dataset, dubbed \textsc{INSPIRED}, and experimentally showed that it can greatly increase the parse accuracy of an initial parser. Moreover, we designed a simulation pipeline to explore the potential of our framework for a variety of semantic parsers, without further annotation effort. The performance improvement shows interactive semantic parsing is promising for further improving KBQA in general.

\section{Ethical Considerations}

\textbf{IRB Approval.} Prior to collection of the INSPIRED dataset, we obtain IRB (Institutional Review Board) approval at our institution. This data collection is considered Exempt Research, meaning that our human subjects are presented with no greater than minimal risk by their participation. Participants' personal information is not collected, aside from minimal demographic information including their native language, which is used to ensure native-speaker level proficiency in the dataset. No identifying information is included. Further, all participants are required to read and agree to an \textit{informed consent} form before proceeding with the task. AMT automatically anonymizes crowdworkers' identities as well. 

\noindent
\textbf{Compensation to Crowdworkers.} In order to ensure both quality data collection and fair treatment of our crowdworkers, we carefully review our payment plan for the AMT task. After a pilot study we gauge the average amount of time we expect a task to require and adjust the payment amount per task according to the minimum wage amount in our state, resulting in a 70 cent payment per task. Further, we ensure compensation for the time spent on the tutorial and qualification task by awarding \$10 bonuses after completion of their first 10 tasks. They also receive \$10 bonuses upon every 100 tasks they complete. In total, the cost of creating the INSPIRED dataset is approximately \$13,300.

\section*{Acknowledgments}

We thank the Clippers and Pragmatics groups at OSU for helpful discussion. We also thank Yu Su, Xiang Deng, Luke Song and Vardaan Pahuja for their valuable feedback. This research was partially supported by a collaborative open science research agreement between Facebook and The Ohio State University. MW has been a paid consultant for Facebook during the period of research. HS was partly supported by NSF IIS-1815674 and NSF CAREER \#1942980.

\bibliography{anthology}
\bibliographystyle{acl_natbib}

\appendix

\section{Dataset Creation Details}
\label{sec:appendix}

The creation of the \textsc{INSPIRED} dataset requires careful selection of questions, design of a decompositional approach, and a translation strategy between logical forms and human-readable language. Further, we carefully design a crowdsourcing task to gather more natural-sounding questions to enhance the quality and versatility of our framework. 

\subsection{Forming Dialogues from CWQ}
\label{AP1_selection_decomposition}

We utilize the \textsc{ComplexWebQuestions} 1.1 (CWQ) dataset \cite{talmor2018repartitioning, talmor2018web}, as this is a common dataset used for complex question-answering over knowledge bases. This dataset is formed by combining questions from the \textsc{WebQuestionsSP} dataset \cite{yih2016value} to form multi-hop complex questions, meaning that they require more than one step to answer. Each question has an associated SPARQL query that functions as a meaning representation of the question. Table \ref{tab:CWQ_example} shows an example of a complex question, its associated SPARQL query, and its answer.

% \hs{see the previous comment; the logic here seems not right. Your purpose is not to study decompositional, right? Decomposition of the logical form is your strategy to explain to the user}

%\st{To create a dialogue around this decomposition process,}\hs{again, the decomposition process is not the goal, but your strategy to explain a logical form to the user, right?} 

\begin{table}[t!]
\begin{center}
\resizebox{0.9\linewidth}{!}{
\begin{tabular}{|c|}
\toprule
\hline
\textbf{Question} \\
\makecell{What is the official language of the country\\that contains Al Sharqia Governorate?
}\\
\hline
\textbf{SPARQL Query} \\
\makecell{<sparql-header-1> ?c ns:location.country.\\administrative\_divisions \#entity1\# . ?c\\ns:location.country.official\_language ?x .}  \\
\hline
\textbf{Answer} \\
Modern Standard Arabic \\
\hline
\bottomrule
\end{tabular}
}
%\vspace{-5pt}
\caption{Example question from the CWQ dataset. The entity ``Al Sharqia Governorate'' is replaced with ``\#entity1\#''. Entities are delexicalized in order to increase generalizability across questions in training.}

\label{tab:CWQ_example}
\end{center}
%\vspace{-15pt}
\end{table}

We envision that a human user will ask a complex question, the system will predict a SPARQL query for that question, decompose it into pieces, translate those pieces into English to show to the user to solicit feedback. The system will then use that feedback to correct the initial parse, if necessary. Figure \ref{fig:dialog_example_1} shows illustrations of this process. 

In order to model this type of dialogue, we utilize a transformer-based seq2seq model~\cite{vaswani2017attention} to predict a SPARQL query for each complex question and decompose the predicted and gold query into pieces, then use these pieces as editable chunks which can be deleted, replaced, or inserted to transform the predicted query into the gold. This process is the framework around which each dialogue is constructed. We translate each step from SPARQL into English to be comprehensible to a human user, thus resulting in dialogues like the one shown in Figure \ref{fig:dialog_example_1}, all stemming from questions that occur in the CWQ dataset. Note that the parser used for this purpose is not state-of-the-art, as part of the goal is to have a broad coverage of error types for correction.

\subsection{Translation of SPARQL Using Templates}
\label{DC_translation}

\begin{table}
\resizebox{1\linewidth}{!}{
\begin{tabular}{|c|l|}
\toprule
\hline
\multicolumn{2}{|c|}{\textbf{Composition}} \\
\hline
Question & \makecell[l]{What is the mascot of the team that has\\Nicholas S. Zeppos as its leader?} \\
\hline
SPARQL & \makecell[l]{<sparql-header-1> ?c ns:organization.\\organization.leadership ?k . ?k\\ ns:organization.leadership.person\\ \#entity1\# . ?c ns:education.educational\\\_institution.mascot ?x .} \\
\hline
Templates & \makecell[l]{1. the organization whose leadership\\includes a person named <PH>} \\
 & \makecell[l]{2. the educational institution with\\the mascot <PH>} \\
\hline
Translation & \makecell[l]{1. \textbf{What is/are} the organization whose\\leadership includes a person named\\\textbf{Nicholas S. Zeppos}?}\\
 & \makecell[l]{2. \textbf{That entity is/are} the educational\\institution with the mascot \textbf{what}?}\\
\hline
\multicolumn{2}{|c|}{\textbf{Conjunction}} \\
\hline
Question & \makecell[l]{What country with the capital of\\Hagåtña is where Sam Shepard lives?} \\
\hline
SPARQL & \makecell[l]{<sparql-header-2> \#entity1\#\\ns:people.person.places\_lived ?y . ?y\\ns:people.place\_lived.location ?x . ?x\\ns:location.country.capital \#entity2\# .} \\
\hline
Templates & \makecell[l]{1. the person(s) who lived in <PH>} \\
 & \makecell[l]{2. the location with the capital city\\named <PH>} \\
\hline
Translation & \makecell[l]{1. \textbf{Sam Shepard is/are} the person(s)\\who lived in \textbf{what}?}\\
 & \makecell[l]{2.\textbf{Of which, what is/are} the location\\with the capital city named \textbf{Hagåtña?}}\\
\hline
\multicolumn{2}{|c|}{\textbf{Comparative}} \\
\hline
Question & \makecell[l]{What country is in the Caribbean with a\\country calling code higher than 590?} \\
\hline
SPARQL & \makecell[l]{<sparql-header-2> \#entity1\#\\ns:location.location.contains ?x . ?x\\ns:common.topic.notable\_types \#entity2\#\\. ?x ns:location.country.calling\_code\\ ?num . filter ( xsd:integer ( ?num ) > 590 ) .} \\
\hline
Templates & \makecell[l]{1. the location(s) containing <PH> (<RSTR>)} \\
 & \makecell[l]{2. the country/countries whose calling\\code is/are <PH>} \\
\hline
Translation & \makecell[l]{1. \textbf{Caribbean is/are} the location(s)\\containing \textbf{what (country)}?}\\
 & \makecell[l]{2. \textbf{Of which, what is/are} the country/\\countries whose calling code is/are\\\textbf{greater than 590}?}\\
\hline
\multicolumn{2}{|c|}{\textbf{Superlative}} \\
\hline
Question & \makecell[l]{Which pro athlete started his career earliest\\and was drafted by the Cleveland Browns?} \\
\hline
SPARQL & \makecell[l]{<sparql-header-2> \#entity1\#ns:sports.\\professional\_sports\_team.draft\_picks\\?y . ?y ns:sports.sports\_league\_draft\_pick.\\player ?x . ?x ns:sports.pro\_athlete.\\career\_start ?num . \} order by ?num limit 1} \\
\hline
Templates & \makecell[l]{1. the team(s) that drafted the athlete(s) <PH>} \\
 & \makecell[l]{2. the pro athlete\lm{(s)} who started their career\lm{(s)} in\\<NUM>} \\
\hline
Translation & \makecell[l]{1. \textbf{Cleveland Browns is/are} the team(s) that\\drafted the athlete(s) \textbf{what}?}\\
 & \makecell[l]{2. \textbf{These entities are} the pro athlete(s) who\\started their career(s) in \textbf{what}?}\\
 & \makecell[l]{3. \textbf{Of these, which is the entity associated}\\\textbf{with the earliest date?}}\\
\hline
\bottomrule
\end{tabular}}
%\vspace{-5pt}
\caption{Question types from the CWQ dataset and the translation process to templated sub-questions.}
\label{tab:question_type}
%\vspace{-15pt}
\end{table}

As this dataset leverages SPARQL queries, we then develop a strategy for how to represent these queries in a more comprehensible form that humans can understand. Thus we create a template corpus and develop a rule-based translation method to do so. The corpus consists of 772 different predicates that appear in the CWQ dataset and translations of each into a basic template that conveys the content. The strategy of using templates to make content more human-friendly has a long history, both utilizing handcrafted templates \cite{Kukich1983DesignOA, mckeown1985discourse, mcroy2000yag} and rule-based template formation \cite{angeli2010simple, kondadadi2013statistical}. We use a blend of both approaches to create templates to represent logical forms in a way that is understandable to our crowdworkers. As can be seen in Table \ref{tab:CWQ_example}, SPARQL queries contain predicates that appear in the form of triples with each component separated by periods, such as \textit{location.country.administrative\_divisions} and \textit{location.country.official\_language}. These triples consist of a domain (\textit{location}), a type (\textit{country}) that represents a class within the domain, and a property (\textit{administrative\_divisions} and \textit{official\_language}) that specify more granular information. These predicates represent content information about the question and can appear in multiple, different questions. For example, the \textit{location.country.administrative\_divisions} predicate maps to the template \textit{the country/countries that contain(s) <PH>}, where <PH> (``placeholder'') gets replaced with a specific entity.

In the parsing process, we delexicalize these specific entities in order to make questions more generalizable and reduce noise during training. For example, in the SPARQL query in Table \ref{tab:CWQ_example}, the replacement token \textit{\#entity1\#} appears, which we replace with \textit{Al Sharqia Governorate} when the template is invoked.

The remaining components of the SPARQL query specify the question type and any additional components, which we leverage to transform the template into a full sentence. The components will be discussed more fully in \ref{DC_types}.  Thus, this particular SPARQL query translates to the following sub-questions:

\begin{enumerate}
    \item \textit{What is/are the country/countries that contain(s) [Al Sharqia Governorate]?}
    
    \textit{ANSWER: Egypt}
    
    \item \textit{That entity is/are the country/countries whose official language is what?}
    
    \textit{ANSWER: Modern Standard Arabic}
\end{enumerate}

\subsection{Question Types}
\label{DC_types}
Each of the questions in CWQ can be categorized into one of four major reasoning types: composition, conjunction, comparative, and superlative \cite{talmor2018web}. Each type can be identified by the SPARQL query and translated accordingly. Table \ref{tab:question_type} shows the translation process of the four types with examples of each. The general strategy is to append content to the beginning of the template and replace the <PH> token to form a complete question and express the appropriate question type. As seen in Table \ref{tab:question_type}, this is quite straightforward for composition- and conjunction-type questions.
 
\textbf{Composition questions} are composed of two simple questions, where the answer to the first is used to form the second question. As an example, in order to answer the question \textit{What is the mascot of the team that has Nicholas S. Zeppos as its leader?}, one must first answer \textit{In which organization is Nicholas S. Zeppos a leader?} to have all the content necessary to answer \textit{What is the mascot of that organization?}. To translate these question types to templated sub-questions, we simply append \textit{What is/are} before the first template and insert the named entity where the <PH> token appears in the template. Then, \textit{That entity is/are} is appended to the beginning of the second template and \textit{what} replaces the placeholder. Note that these positions can be reversed depending on what content is provided in the question. For example, a question could be either of the two options, depending on the goal of the target question:

\begin{enumerate}
    \item \textit{\textbf{What is/are} the organization whose leadership includes a person named \textbf{Nicholas S. Zeppos}?}
    \item \textit{\textbf{Vanderbilt University is/are} the organization whose leadership includes a person named \textbf{what}?}
\end{enumerate}

\textbf{Conjunction questions} follow a very similar process, though because their goal is to find the intersection of two categories, the first question returns a list of answers. To account for this, we simply append \textit{Of which} to the second question before following the same set of rules as the composition questions. 

\textbf{Comparative questions} generally have a comparative operator (<, >) and a number contained in their SPARQL query, which we translate simply to \textit{less than X} or \textit{greater than X}, as appropriate. Note that the comparative example in Table \ref{tab:question_type} contains a ``restriction predicate'', marked by the <RSTR> token. This will be discussed in Section \ref{DC_LF_features}.

\textbf{Superlative questions} require a slightly more complicated strategy. The first sub-question of a superlative type question always generates a list of answer options, while the second sub-question must pair those answer options with numerical information, such as dates or integers. Then, these numbers are ordered, either from smallest-to-largest or vice versa, and the first is returned as the final answer. To account for this, we append \textit{These entities are} to front of the second template, to make it clear that multiple entities are involved, and return a paired list of entities and their corresponding values as an answer. Then we append a third sub-question that specifies how the questions are sorted and returns a single answer.

\subsubsection{Logical Form Features}
\label{DC_LF_features}

Within the four main types of questions (composition, conjunction, comparative, and superlative), there are a variety of features that appear. These features include filters, restriction predicates, and union predicates.

\textbf{Filters} act to restrict a list of entities in some fashion by assigning numerical boundaries. An example of this can be seen in Table \ref{tab:question_type} in the comparative question's SPARQL query, starting with the word \textit{filter}. This sequence limits the list of entities by ones whose calling codes are larger than 590.

\textbf{Restriction predicates} can appear as auxiliary pieces to regular predicates and typically provide categorical information about an entity. For example, in Table \ref{tab:question_type}, the comparative-type question \textit{What country is in the Caribbean with a country calling code higher than 590?} has two entities in its SPARQL query, though \textit{Caribbean} is the only entity that seems to appear in the original question. The two main predicates are \textit{location.location.contains} and \textit{location.country.calling\_code}, but a third predicate, \textit{common.topic.notable\_types} appears in between them. This predicate acts as a restriction upon the first main predicate; in this case \textit{\#entity2\#} corresponds to \textit{country} and restricts the locations that can appear as answers to the category of countries. 

Because restriction predicates are not stand-alone pieces that could be translated into their own sub-questions, we develop a strategy for incorporating them into the templates of the predicates they restrict. First, we create a corpus of ``mini-templates'' that correspond to all the restriction predicates that could appear. Much of the time, these mini-templates simply place the entity (like \textit{country} in the previous example) into parentheses, though in some cases they situate the entity into a prepositional phrase. 

Meanwhile, the main template corpus has tokens in place to define where the mini-template should be placed in the main template. One can see in the comparative example of Table \ref{tab:question_type} that there is an <RSTR> token in the template of the first sub-question. Every main template that can appear with a restriction predicate has this token in its template; though it needs not always appear with one. Consequently, if the restriction token does not get replaced, it simply gets deleted. If the \textit{location.location.contains} predicate appeared without a restriction predicate, it would simply read \textit{Caribbean is/are the location(s) containing what?}

\begin{table}
\resizebox{1\linewidth}{!}{
\begin{tabular}{|c|l|}
\toprule
\hline
\multicolumn{2}{|c|}{\textbf{Composition}} \\
\hline
Question & \makecell[l]{Who is both a member of the Kennedy\\family and the Order of the British Empire?} \\
\hline
SPARQL & \makecell[l]{\textbf{filter ( ?x != \#entity1\# ) \{ \# parents \#entity2\# ns:}\\\textbf{people.person.parents ?x . \} union \{ \# children}\\\textbf{\#entity3\# ns:people.person.children ?x .}\\\textbf{\} union \{ \# siblings \#entity4\#  ns:people.person.}\\\textbf{sibling\_s ?y . ?y ns:people.sibling\_relationship.}\\\textbf{sibling ?x . \} union \{ \#spouse \#entity5\# ns:}\\\textbf{people.person.spouse\_s ?y . ?y ns:people.}\\\textbf{marriage.spouse ?x . ?y ns:people.marriage.}\\\textbf{type\_of\_union \#entity6\# . filter ( not exists \{ ?y}\\\textbf{ns:people.marriage.to []\}) \} }\\?x ns:royalty.chivalric\_order\_member.belongs\_\\to\_order ?c . ?c ns:royalty.chivalric\_order\_\\membership.order \#entity7\# .} \\
\hline
Templates & \makecell[l]{1. the family of <PH>} \\
 & \makecell[l]{2. the member(s) of the order of <PH>} \\
\hline
Translation & \makecell[l]{1. \textbf{Who is/was} the family of \textbf{John F. Kennedy}?}\\
 & \makecell[l]{2. \textbf{Of which, what is/are} the member(s) of the \\order of \textbf{Order of the British Empire}?}\\

\hline
\bottomrule
\end{tabular}}
%\vspace{-5pt}
\caption{Example of a question whose SPARQL query includes a union predicate.}
\label{tab:union_predicates}
%\vspace{-15pt}
\end{table}

\textbf{Union predicates} are a bit of a misnomer, as they are actually a group of predicates that function as though they are a single predicate, and thus correspond to a single template. In Table~\ref{tab:union_predicates}, one can see that the SPARQL query is quite long, with all of the content in bold corresponding to the first sub-question and the remainder corresponding to the second. Within this first sub-LF, there are several predicates that are joined together by \textit{\} union \{}. Collectively, these templates encompass the concept of \textit{family} by defining all the various relationship roles that are involved in that concept. Theoretically, we could enumerate all of these in template form, separated by \textit{or} (\textit{the brother of John F. Kennedy \textbf{or} the mother of John F. Kennedy \textbf{or} the child of John F. Kennedy}...) but this seems to be an unnecessarily complicated and inconcise way of representing these. Instead, we enumerate the various types of union predicates that could appear and create a small corpus of templates that express the overall concept represented by each collection of predicates, thus crowdworkers will see questions with this feature in the same format as a regular question.

\subsection{Crowdsourced Data Collection}
\label{AP1_crowdsource}

As mentioned in Section \ref{crowdsourcing}, the crowdsourcing task for this dataset is primarily a paraphrasing task in which crowdworkers work through a structured dialogue, rephrasing templated sub-questions at each step. 

Each task takes the form of a dialogue involving three entities: the ``user'', which is an automated dialogue partner, an automated ``director'' that guides the dialogue and provides detailed instructions, and the ``agent'', which is the role performed by the crowdworker. Upon entering a task, the worker is shown the ``target question'', or the original question from CWQ, and asked if the question was sensible to them. If so, they are asked to rephrase it using different language. If not, they proceed with the dialogue in the hopes that the decomposition process will make the meaning of the question clear. In these cases, the crowdworker is asked to rephrase the target question at the end of the dialogue. This process is included to encourage better understanding of the target question and to help us recognize confusing questions in the original dataset and replace them with higher-quality questions when appropriate.

\begin{figure*}[!t]
    \centering
    
    \includegraphics[width=0.7\linewidth]{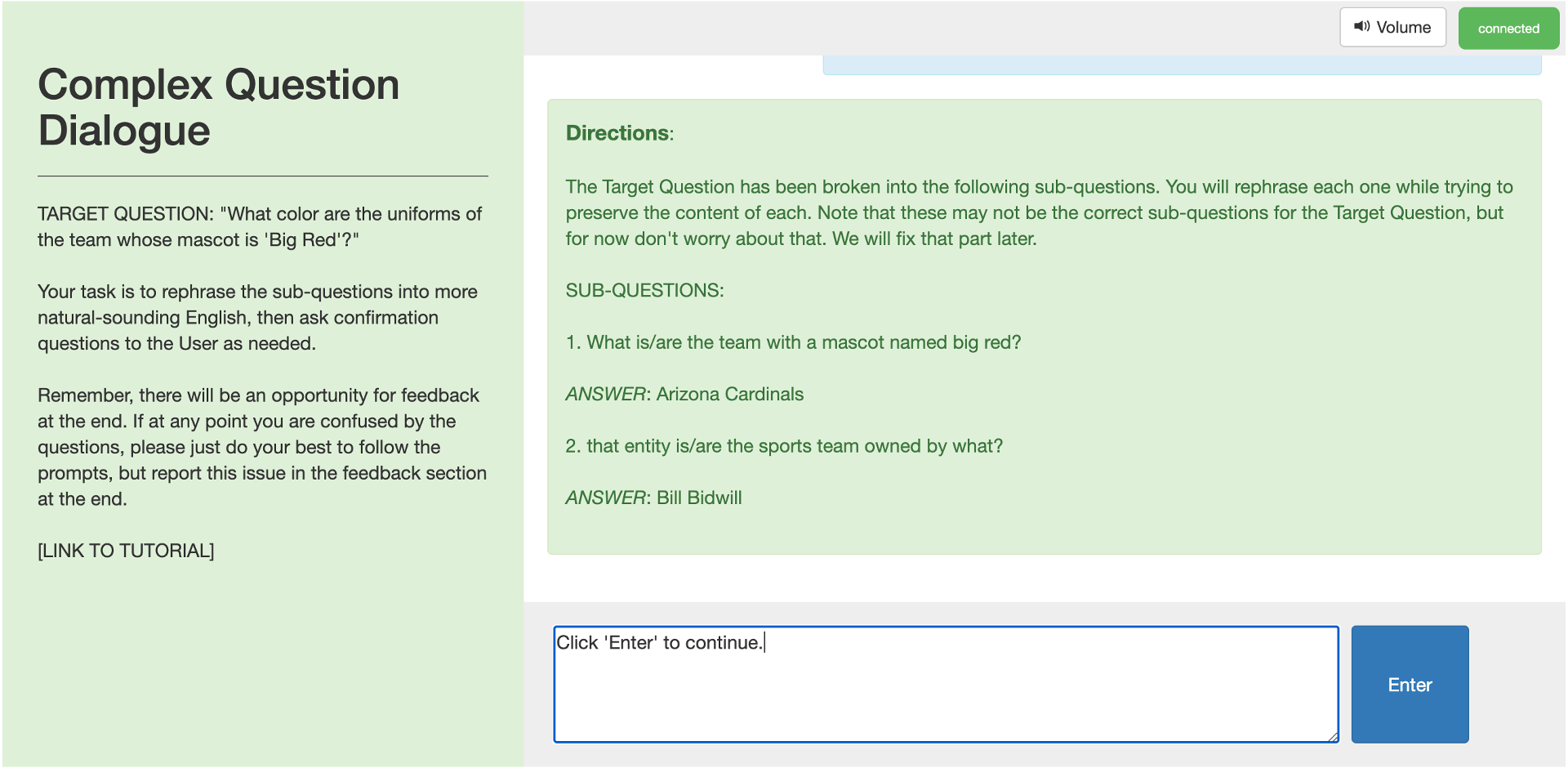}
    %\vspace{-5pt}
    \label{fig:parlai_1}

    \includegraphics[width=0.7\linewidth]{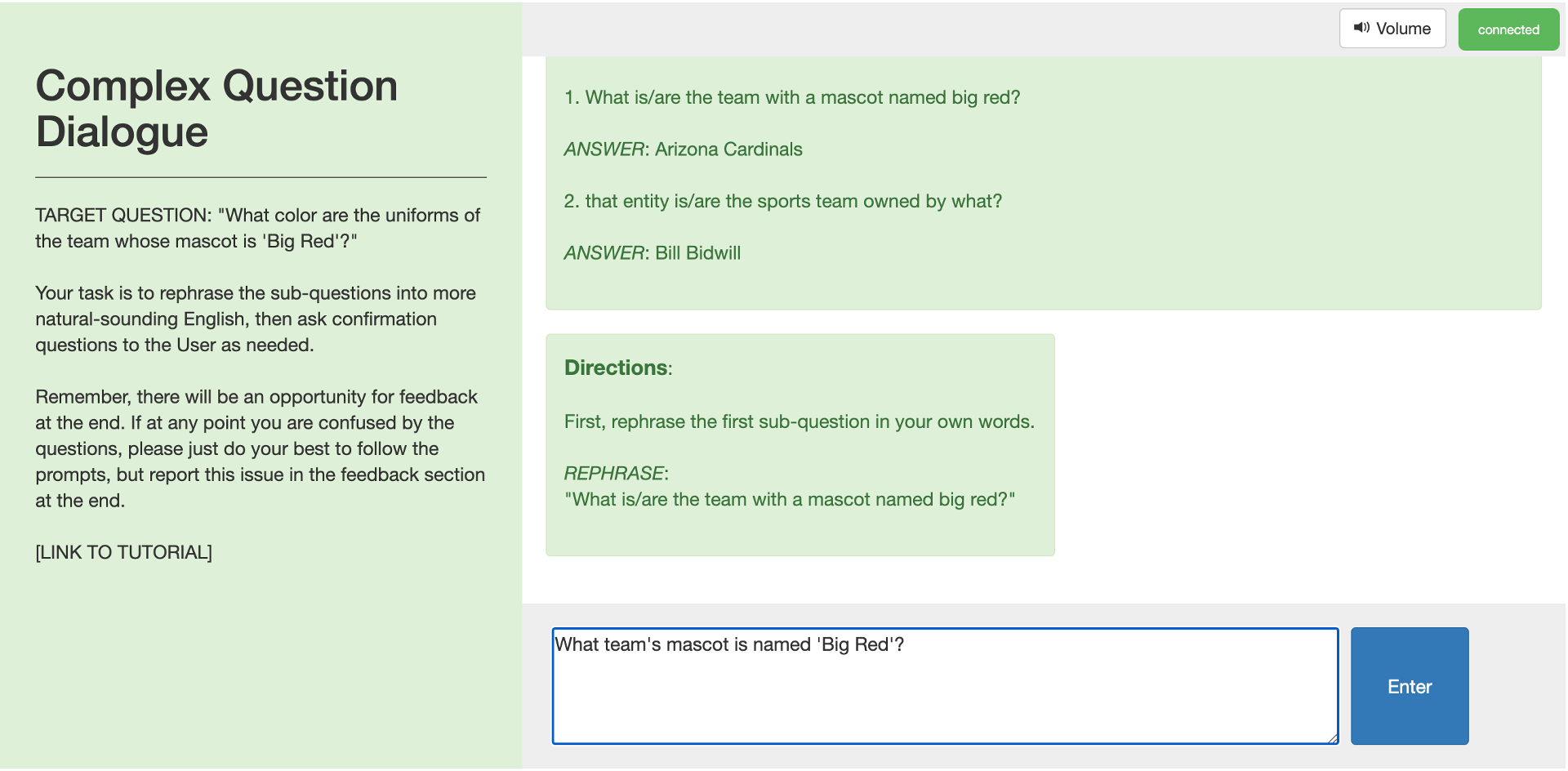}
    %\vspace{-5pt}
    \caption{Data collection interface on AMT, using the ParlAI framework \cite{miller2017parlai}.}
    \label{fig:parlai_2}

\end{figure*}

Next, the target question is automatically decomposed into templated sub-questions which are displayed to the worker, who rephrases them into English. These rephrased questions are sent to the automated user, who provides corrections as necessary. The worker rephrases any new questions and the edits are automatically made. At the end of the dialogue, the worker is asked for any feedback regarding the dialogue. This feedback is later used to make corrections and flag any problems that might have arisen. Screenshots of the dialogue interface can be seen in Figure \ref{fig:parlai_2}.

\section{Dataset Analysis}
\label{sec:appendix3}

\subsection{Cleaning the Dataset}
\label{AP1_data_cleaning}

As mentioned in Section \ref{dataset_analysis}, we employ a semi-automatic data cleaning method to reduce the error rate in the \textsc{INSPIRED} dataset. Because data cleaning can be an expensive and time-consuming process, the goal is to develop a method that would reduce the number of items in the dataset that need to be manually reviewed. Thus we use an automatic method to identify a small subset of the entire dataset that contain as many errors as possible to then manually review. To this end, we utilize a pretrained sequence-to-sequence model that employs the idea of \textsl{cycle consistency} \cite{Zhu_2017_ICCV}, to identify poor paraphrases by retrieving meaning representations (MRs) from questions rephrased by the workers. Then these MRs are used to compare against the original MRs and evaluated for similarity.  

In order to evaluate the effectiveness of the strategy, a random 5\% subset of the entire dataset is selected for annotation, using a binary classification of whether or not the rephrased question was an accurate paraphrase of the original templated question (and by extension, its original logical form). This annotation effort revealed that 4.4\% of the rephrased questions contain errors, which we expect is representative of the entire dataset. 

We then fine-tune Hugging Face's implementation of T5 in a seq2seq model to generate MRs, in this case templated sub-questions, to compare to the original MRs \cite{wolf-etal-2020-transformers, raffel2020exploring}. These pairs of MRs then need to be sorted in a ranked list that filters paraphrases that are more likely to contain errors to the top of the list. This allows us to use a \textsl{precision at K} measure, which, given a rank \textsl{K}, the precision is calculated over the set of retrieved items with a rank of \textsl{K} or less. For the annotated test set, \textsl{K} equals 75, the number of observed errors. After ranking the list, we can evaluate the quality of the method by checking the top \textsl{K} data points and checking to find how many errors appear in that set, compared to a random baseline of 4.4\% (the observed error rate), or about 3 errors. 

We employ two ranking methods to sort the pairs. First, we calculate the negative log-likelihood of the target MRs relative to the model and then do the same for the generated MRs.

\begin{equation}
\label{calc_score}
\mathcal{S}(y) = -\sum\limits_{y_i\in Y}\log p(y_i | y_{<i},x;\theta)
\end{equation}

\begin{center}
$y = \langle y_1, ... , y_{|y|}\rangle$

$y_{<i} = \langle y_1, ... , y_{i-1}\rangle$
\end{center}

$S(y)$ refers to the score of a given output sequence $y$, which is the sum of the negative log-likelihood of each $y_i$ given the sequence of $y$ tokens that come before. $\theta$ refers to the model parameters.

Once the negative log-likelihoods are determined for each candidate $y$, the best candidate is determined based on the lowest score. 

\begin{equation}
y* = argmin (S(y|x))
\end{equation}

Here, $y*$ refers to the best generated output sequence, and $x$ is a given input sequence. A score for output sequence $y*$ is determined, as well as a score for the target sequence $t$.

\begin{equation}
\label{diff_eq}
D = |S(y*) - S(t)|
\end{equation}

While these two scores are comparable to each other, they are not comparable across other item pairs. In order to assign a ranking for every item in the dataset, we calculate the difference $D$ between the negative log-likelihoods of the target MR and generated MR for each question in the dataset and sort them based on the largest difference score, as shown in Equation \ref{diff_eq}.

Second, we calculate an edit distance score between the target MR and generated MR and sort based on the largest score. If the model has predicted an MR that is substantially far from the target MR in its phrasing, it likely has a different meaning. 

Using the first ranking method, 17.3\% of the errors are recovered, while the second recovers 32\% of the errors. However, because the two ranking methods appear to be identifying different errors with little overlap, both are used to identify the final set of questions for manual review, drawing from the methods equally.

Then the method is applied to the entire \textsc{INSPIRED} dataset, using cross-validation with a series of 90\% training, 10\% testing splits to generate MRs for every rephrased question. Then, because the annotated dataset has a 4.4\% error rate which we expect to be representative of all the data, the top-ranked ~4.4\% of data is selected for manual review. This review results in 17.7\% of items being revised, meaning that authors change the rephrasing to more accurately reflect the original meaning.

\subsection{GEM Metrics}
\label{sec:appendix2}
\begin{table}[H]
\begin{center}
\resizebox{1\linewidth}{!}{
\begin{tabular}{ccc}
\toprule

 & \textbf{Template Corpus} & \textbf{Rephrased Corpus} \\
 \hline
\textbf{Unigrams} & \\
 \hline
\makecell[r]{Vocab Size} & 8,465 & 9,864 \\
\makecell[r]{Distinct} & 0.012 & 0.022 \\
\makecell[r]{Unique} & 1,003 & 1,258 \\
\makecell[r]{Entropy} & 6.532 & 8.090 \\
 \hline

\textbf{Bigrams} & \\
 \hline
\makecell[r]{Vocab Size} & 21,072 & 44,085 \\
\makecell[r]{Distinct} & 0.031 & 0.109 \\
\makecell[r]{Unique} & 2,949 & 8,723 \\
\makecell[r]{Entropy} & 8.976 & 12.484 \\
\makecell[r]{Cond Entropy} & 2.295 & 3.918 \\
 \hline
\textbf{Trigrams} & \\
 \hline
\makecell[r]{Vocab Size} & 31,838 & 81,479 \\
\makecell[r]{Distinct} & 0.050 & 0.224 \\
\makecell[r]{Unique} & 5,332 & 20,971 \\
\makecell[r]{Entropy} & 10.291 & 14.529 \\
\makecell[r]{Cond Entropy} & 1.250 & 1.986 \\

\bottomrule
\end{tabular}
}
%\vspace{-5pt}
\caption{GEM n-gram metrics for the template corpus and rephrased question corpus.}

\label{tab:APPENDIX_diversity_metrics}
\end{center}

\end{table}

Table \ref{tab:APPENDIX_diversity_metrics} shows the N-gram statistics of all the templates in the dataset (template corpus) and all the rephrased questions (rephrased corpus). These metrics are calculated using the GEM evaluation scripts \cite{gehrmann2021gem}. In this table, \textsl{Vocab Size} refers to the total number of distinct N-grams, while \textsl{Distinct} refers to the ratio of distinct N-grams divided by the total number of N-grams in the dataset. \textsl{Unique} specifies the number of N-grams that occur only once in the dataset, \textsl{Entropy} is the Shannon entropy over N-grams, and \textsl{Cond(itional) Entropy} is the entropy conditioned on N$_{\minus 1}$-grams.

\subsection{Lexical Analysis}
\label{sec:lexical}
\begin{table}[t!]
\begin{center}
\resizebox{0.8\linewidth}{!}{
\begin{tabular}{l|c}
\toprule
\textbf{Lexical Relationship} &  \textbf{Percentage(\%)} \\
\midrule
Lexical Match  &  58 \\
\midrule
Synonymy & 31 \\
Hypernymy & 5 \\
Hyponymy & 20 \\
\bottomrule
\end{tabular}}

\caption{Lexical analysis of 100 randomly sampled sub-questions and their templates. Note that \textsl{Lexical Match} refers to the percentage of words in all sub-questions that appear in their corresponding templated question.}
\label{tab:paraphrase_characteristics}
\end{center}

\end{table}

In order to better understand the methods by which crowdworkers rephrased templates, 100 randomly selected sub-questions are studied in terms of the lexical relationships between the template and rephrased versions. Table \ref{tab:paraphrase_characteristics} shows the results of this analysis. ``Lexical match'' refers to the average proportion of words in the rephrased version that also appear in the template, relative to the total number of words in the rephrased version. Synonymy, hypernymy, and hyponymy refer to the number of questions in the 100 selected items that contain an instance of one of these lexical relations. It is clear, therefore, that crowdworkers are using these strategies in their rephrasings of the templates, in addition to simply changing word order. On average, a bit less than half the words in a rephrased question are newly introduced by the crowdworker, and 56\% of the time they are using synonmy, hypernymy, hyponymy, or some combination of these to rephrase the templated question.

\begin{table*}[t!]
\begin{center}
\resizebox{1\linewidth}{!}{
\begin{tabular}{c|c|c}
\toprule

\makecell[l]{\textbf{Sub-question predicate}} & \textbf{Actual Context} & \textbf{Random Context}  \\
\hline

\makecell[l]{film.film\_subject.films} & \makecell[l]{Complex Question: \textbf{Who was} the wife of \textbf{the subject of} \\\hspace{3.1cm} \textbf{the} film \#entity\#? \\ Sub-Questions: \\ \hspace{.35cm} *1. Who was the subject of the movie \#entity\#? \\ \hspace{.5cm} 2. Who was that person married to?} & \makecell[l]{Complex Question: Where did the topic \textbf{of the} film \#entity\# \\\hspace{3.1cm} pass away at?\\ Sub-Questions: \\ \hspace{.35cm} *1. Who was the main focus in the movie called \#entity\#? \\ \hspace{.5cm} 2. Where did this individual die?}\\

\hline

\makecell[l]{influence.influence\_node.\\influenced\_by} & \makecell[l]{Complex Question: Which peer of \#entity\# \textbf{inspired} \\ \hspace{3.1cm} \textbf{the work of} \#entity\#? \\ Sub-Questions: \\ \hspace{.35cm} *1. Who inspired the work of \#entity\#? \\ \hspace{.5cm} 2. Of the above named people, which had a peer \\ \hspace{.95cm} relationship with \#entity\#?} & \makecell[l]{Complex Question: What person \textbf{who} influenced \#entity\# 's \\ \hspace{3.1cm} \textbf{work} was born on \#entity\#? \\ Sub-Questions: \\ \hspace{.35cm} *1. \#entity\# inspired which people's work? \\ \hspace{.5cm} 2. Which of these people were born on \#entity\#?}\\

\bottomrule
\end{tabular}
}
%\vspace{-5pt}
\caption{Examples of sub-questions in their actual context vs.\ a random context that utilizes the same predicate in its logical form. The sub-question was substituted for the one that used the same logical form (marked with *) in the random context when calculating ROUGE scores. Lexical overlap of the sub-question with each context is represented by bold text. Entities have been replaced with \#entity\# tokens in order to avoid disadvantaging the random context due to overlap in named entities.}

\label{tab:ROUGE_examples}
\end{center}
%\vspace{-15pt}
\end{table*}

\subsection{Contextual Awareness}
\label{contextual_awareness}

In a given dialogue, we provide answers to the sub-questions when possible, making the dialogue context-rich and providing the user with as much information as possible to help them understand the decomposition process of their original query.

This context-awareness can also be seen in the sub-question paraphrases. Our crowdworkers are encouraged to paraphrase questions in a manner that accounts for the overall context of the question, particularly with regard to named entities. For example, when a second sub-question references the answer of the first sub-question, we ask the Turkers to reference that entity without naming it explicitly, but also using a more specific phrase than \textit{entity}. An example of this can be seen in Figure \ref{fig:dialog_example_1}, where instead of directly incorporating the answer of the first question (\textit{Egypt}) into the second question, they reference it using the phrase \textit{the above-named nation}. The goal of this strategy is to create a dataset of dialogues that are context-aware and grounded, on which generation models can be trained to mirror this behavior. \al{By using less specific phrases than entity names, our model is better able to generalize across examples during training.} 

However, one can envision that in a real-use situation, it might be more natural for a user to simply use \textit{Egypt} instead of \textit{the above-named nation} when correcting sub-question 2. While our current framework is not able to accommodate this behavior, a simple data augmentation procedure in which referring expressions are replaced with the named entities should allow our system to accommodate this. We leave this data augmentation for future work, but plan to implement it upon conducting a study with real users. 

In order to demonstrate contextual awareness, Table \ref{tab:context_aware} in Section~\ref{dataset_analysis} shows the average ROUGE-1 and ROUGE-2 scores of all sub-questions in their actual contexts compared with the same sub-questions in a randomly assigned context that utilizes the same sub-logical form. The higher scores for the actual context indicate that the wording of sub-questions reflect the context from which they are derived. Moreover, Table \ref{tab:ROUGE_examples} shows examples of sub-question with these context comparisons.

\subsection{Error Characteristics of Initial Parser}
\label{error_char}

\begin{table}[t!]
\large
\begin{center}
\resizebox{1\linewidth}{!}{
\begin{tabular}{rcccc|c}
\toprule
  & \textbf{Conjunction} & \textbf{Composition} & \textbf{Comparative} & \textbf{Superlative} & \textbf{Total}\\
\textbf{Delete} & 49 & 94 & 9 & 3 & 155 \\
\textbf{Insert} & 1835 &765 & 208 & 208 & 3016 \\
\textbf{Replace} & 2207 & 1757 & 341 & 393 & 4698 \\
\textbf{No Action} & 2172 & 2240 & 301 & 310 & 5023 \\

\bottomrule
\end{tabular}
}
%\vspace{-5pt}
\caption{Distribution of error types within the sub-questions of the four main question types.}

\label{tab:edit_operations_by_type}
\end{center}
%\vspace{-15pt}
\end{table}

It is important to note that our initial parser is purposefully not state-of-the-art, as we want to have a wide distribution of errors around which we could create dialogues. \mww{(See Section~\ref{sec:experiments} for details about the initial parser.)} Similar to other interactive semantic parsing work, we envision that the user will provide corrections to the sub-questions, though we at this stage require the user to use the three operations of deleting, replacing, or inserting a whole sub-question. Table \ref{tab:edit_operations_by_type} shows the distribution of sub-questions whose original complex question is of each of the four main types. Within these types, the distribution of edit operations per sub-question is shown. Though many of sub-questions do not need any edits, the \textsl{replace} operation is most frequent of edit operations, appearing in roughly 36.5\% of each type, while \textsl{insert} is roughly 23.3\% and \textsl{delete} is around 1.2\%, with \textsl{no action} making up the remaining 39\%. These distributions indicate the parser is more likely to predict something incorrect or leave out a sub-question, rather than predict a sub-question that is not present in the gold.

\section{Sub-Task Error Analyses}
\label{sec:appendix4}

\begin{table}[t!]

\begin{center}
\renewcommand\arraystretch{1.1}

\resizebox{1\linewidth}{!}{
\begin{tabular}{l}
\toprule
\textbf{Sub-Q:} \\
 What tourist attractions are by the grand canyon?\\
\textbf{Gold Sub-LF:} \\
\#entity1\# ns:travel.tourist\_attraction.near\_travel\_destination ?x .\\ 
\textbf{Generated Sub-LF:} \\
\#entity1\# ns:travel.travel\_destination.tourist\_attractions ?x .\\ 
\hline
\textbf{Sub-Q:} \\
What is that country's national anthem? \\ 
\textbf{Gold Sub-LF:} \\
?c ns:location.country.national\_anthem \textbf{?y . ?y} \\ ns:government.national\_anthem\_of\_a\_country.anthem ?x . \\
\textbf{Generated Sub-LF:} \\
?c ns:location.country.national\_anthem ?x . \\

\bottomrule
\end{tabular}
}
%\vspace{-5pt}
\caption{Two error cases about wrongly generated predicates in an analysis of 100 generated sub logical forms. $?y$ in the logical form is an example of CVT node which connects two predicates that operate as a single, compound predicate.}
\label{tab:parse_errors}
\end{center}
\vspace{-10pt}
\end{table}

\textbf{Parse Correction.}  We sample 100 erroneous predictions of BART-large under the best-performing setting in Table~\ref{tab:parse_1_2}. In this analysis, it becomes clear that longer, more complicated logical forms are more likely to be mispredicted. Only 21 of the errors involve single predicates, while 54 erroneous parses occur with \textsl{CVT (Compound Value Type)} predicates, which are essentially two predicates combined together via \textsl{CVT nodes} (for example \textsl{?y} in Table~\ref{tab:parse_errors}) that function as a single predicate. 13 errors occur on \textsl{restriction} predicates, which co-occur with single or \textsl{CVT} predicates to further limit the entity type. For example, predicates of the \textsl{location} domain might occur with a restriction that limits that predicate to locations of the type \textsl{country}. The remaining 12 errors all occur due to only partially generating a long logical form that contains filters. Details regarding restriction predicates and filters can be found in \ref{DC_LF_features}.

\begin{table}[t!]\Huge
\begin{center}
\renewcommand\arraystretch{2.0}
\resizebox{1\linewidth}{!}{
\begin{tabular}{l|l}
\toprule
\fontsize{30}{1.5}\selectfont \textbf{Human-Written} &  \fontsize{35}{1.5}\selectfont \textit{Which of the above named people did the voice of toki?}\\
\fontsize{30}{1.5}\selectfont \textbf{Machine-Generated} & \fontsize{35}{1.5}\selectfont \textit{Which of these people played the role of toki?} \\
\fontsize{30}{3}\selectfont \textbf{Error Explanation} & \fontsize{35}{3}\selectfont \makecell[l]{Generated question does not specify that\\ the role was a voice acting one.} \\
\hline
 \fontsize{30}{1.5}\selectfont \textbf{Human-Written} & \fontsize{35}{1.5}\selectfont \textit{What famous person has addison's disease?} \\
 \fontsize{30}{1.5}\selectfont \textbf{Machine-Generated} & \fontsize{35}{1.5}\selectfont \textit{who has suffer from addison's disease?} \\
 \fontsize{30}{1.5}\selectfont\textbf{Error Explanation} & \fontsize{35}{1.5}\selectfont \makecell[l]{Grammatical error} \\
 \hline
 \fontsize{30}{1.5}\selectfont \textbf{Human-Written} & \fontsize{35}{1.5}\selectfont \makecell[l]{\textit{What district does that politician represent?}} \\
 \fontsize{30}{1.5}\selectfont \textbf{Machine-Generated} & \fontsize{35}{1.5}\selectfont \makecell[l]{\textit{What district does that person represent?}} \\
 \fontsize{30}{1.5}\selectfont \textbf{Error Explanation} & \fontsize{35}{1.5}\selectfont \makecell[l]{Generated question is slightly less specific} \\

\bottomrule
\end{tabular}
}
%\vspace{-5pt}
\caption{Three instances of errors in an analysis of 100 generated sub-questions compared to human-written versions.}
\label{tab:generation_errors}
\end{center}
\vspace{-15pt}
\end{table}

\noindent
\textbf{Sub-Question Generation.} We conduct an analysis on 100 randomly selected pairs of human-written question and machine-generated question that correspond to the same logical form. We first examine questions from the best-performing model in Table~\ref{tab:generation_1_2} according to BLEU scores and BERTScores, which use the current sub-logical form, the current templated sub-question, the complex question and the history of templated sub-questions from previous steps as context. Questions in which the machine-generated and human-written versions exactly match each other were excluded. This analysis reveals that only three generated questions (3\%) are of perceptibly worse quality than the human-written questions, as can be seen in Table~\ref{tab:generation_errors}. Further, there are four cases in which the human-written questions contain grammatical errors, whereas the machine-generated ones do not. An analysis of all generated questions which do not exactly match their human-written counterpart reveals that 64\% of the generated questions are shorter in terms of number of words.

Because BLEU scores do not necessarily paint a full picture of the model performance, we then examine the \mww{generated responses} from the model that produced the lowest BLEU scores, which is the model with no context. By examining the same 100 samples as in the previous analysis, we note twenty cases in which the best-performing model that leverages context better reflects that context in its rephrasing than the model that does not leverage context. There are, however, 6 cases in which the model without context does this better and in the remaining cases there is no discernible difference between the quality of the generations from the two models. Table \ref{tab:generation_errors2} shows examples of each of these cases, for illustration. 

\begin{table}[t!]\Huge
\begin{center}
\renewcommand\arraystretch{2.0}
\resizebox{1\linewidth}{!}{
\begin{tabular}{c|c|c|c}
\toprule

\textbf{Better Model} & \fontsize{30}{1.5}\selectfont Neither & \fontsize{30}{1.5}\selectfont Model 1 & \fontsize{30}{1.5}\selectfont Model 2 \\

\hline

\textbf{Number} & \fontsize{30}{1.5}\selectfont 74 & \fontsize{30}{1.5}\selectfont 20 & \fontsize{30}{1.5}\selectfont 6 \\
\hline
\textbf{EXAMPLES} & \multicolumn{3}{c}{ } \\

\hline

\makecell[c]{Human-\\Written} & \makecell[l]{\fontsize{30}{1.5}\selectfont who were walt\\ \fontsize{30}{1.5}\selectfont disney's kids?} & \makecell[l]{\fontsize{30}{1.5}\selectfont kevin hart went to \\\fontsize{30}{1.5}\selectfont what schools?} & \makecell[l]{\fontsize{30}{1.5}\selectfont what is the name of \\\fontsize{30}{1.5}\selectfont the currency used in \\\fontsize{30}{1.5}\selectfont that country?} \\

\hline

\makecell[c]{Model 1\\(w/context)} & \makecell[l]{\fontsize{30}{1.5}\selectfont who are the children \\\fontsize{30}{1.5}\selectfont of walt disney?} & \makecell[l]{\fontsize{30}{1.5}\selectfont what \textbf{schools} did \\\fontsize{30}{1.5}\selectfont kevin hart go to?} & \makecell[l]{\fontsize{30}{1.5}\selectfont what kind of currency \\\fontsize{30}{1.5}\selectfont do they use?} \\

\hline

\makecell[c]{Model 2\\(w/o context)} & \makecell[l]{\fontsize{30}{1.5}\selectfont what are the names\\ \fontsize{30}{1.5}\selectfont of walt disney's\\ \fontsize{30}{1.5}\selectfont children?} & \makecell[l]{\fontsize{30}{1.5}\selectfont what did kevin \\\fontsize{30}{1.5}\selectfont hart go to?} & \makecell[l]{\fontsize{30}{1.5}\selectfont what is the currency \\\fontsize{30}{1.5}\selectfont used \textbf{in that country}?} \\

\bottomrule
\end{tabular}
}
%\vspace{-5pt}
\caption{Comparison of 100 generated  sub-questions from models with and without context in their inputs. The bolded text in columns 2 and 3 highlight what enhanced the quality of the generation in comparison to its counterpart. Model 1 refers to the model that used the complex question and previous templated questions as context (row 4 in Table \ref{tab:generation_1_2}) and Model 2 refers to the model that did not use context at all (row 1 in Table \ref{tab:generation_1_2}).}
\label{tab:generation_errors2}
\end{center}
\vspace{-15pt}
\end{table}

While these results are based our observations and certainly require further future investigation and human annotation by people other than the authors, these preliminary results show that the generated questions can be more concise and of comparable quality.

\end{document}